\documentclass[letterpaper, 10 pt, conference]{./support/ieeeconf}
\usepackage{times}
\usepackage{amsmath}
\usepackage{stfloats} 
\usepackage[pdftex]{graphicx}
\usepackage{caption}
\usepackage{subfiles} 
\usepackage{subfigure}
\usepackage{graphicx}
\let\labelindent\relax
\usepackage{enumitem}
\usepackage{amsmath,amssymb,amsopn,amstext,amsfonts}
\usepackage{cancel}
\usepackage[space]{cite}
\usepackage{pdfsync}
\usepackage{balance}
\usepackage{color}
\usepackage[table]{xcolor}
\usepackage{times}
\usepackage{algorithm}
\usepackage{algorithmic}
\usepackage{url}
\usepackage{booktabs}
\usepackage{threeparttable}
\usepackage[linkcolor=black,citecolor=black,urlcolor=black,colorlinks=true]{hyperref}
\usepackage{multirow}
\usepackage[table]{xcolor} 
\usepackage{colortbl}
\usepackage[outdir=./epstopdf/]{epstopdf}
\usepackage{graphicx}
\usepackage{array}
\usepackage{verbatim}
\usepackage{makecell}
\usepackage{subcaption}
\usepackage{stackengine}
\usepackage{balance}
\usepackage{paracol}
\IEEEoverridecommandlockouts
\definecolor{mygreen}{RGB}{219, 234, 192}

\graphicspath{{./figure/}}

\DeclareGraphicsExtensions{.pdf,.png,.jpg,.eps,.PNG}
\title{\LARGE \bf MultiPark: Multimodal Parking Transformer with \\Next-Segment Prediction}
\author{Han Zheng$^1$, Zikang Zhou$^2$, Guli Zhang$^2$, Zhepei Wang$^2$, Kaixuan Wang$^2$, \\Peiliang Li$^2$, Shaojie Shen$^3$, Ming Yang$^1$, and Tong Qin$^{1\ast}$ 
    \thanks{
        $^{1}$Shanghai Jiao Tong University, Shanghai, China.
        {\{hanzheng, qintong\}@sjtu.edu.cn}.
    }
    \thanks{
        $^{2}$Zhuoyu Technology, Co., Ltd., Shenzhen, China.
    }
    \thanks{
        $^{3}$Department of Electronic and Computer Engineering, Hong Kong University of Science and Technology, Hong Kong.
    }
    \thanks{
    {$^\ast$ is the Corresponding author. This work was supported by the Natural Science Foundation of Shanghai (Grant No. 24ZR1435600).}
    }
    }

\begin{document}
\maketitle


\begin{abstract}
Parking accurately and safely in highly constrained spaces remains a critical challenge.
Unlike structured driving environments, parking requires executing complex maneuvers such as frequent gear shifts and steering saturation.
Recent attempts to employ imitation learning (IL) for parking have achieved promising results.  
However, existing works ignore the multimodal nature of parking behavior in lane-free open space, failing to derive multiple plausible solutions under the same situation.
Notably, IL-based methods encompass inherent causal confusion, so enabling a neural network to generalize across diverse parking scenarios is particularly difficult.
To address these challenges, we propose MultiPark, an autoregressive transformer for multimodal parking.
To handle paths filled with abrupt turning points, we introduce a data-efficient next-segment prediction paradigm, enabling spatial generalization and temporal extrapolation.
Furthermore, we design learnable parking queries factorized into gear, longitudinal, and lateral components, parallelly decoding diverse parking behaviors.
To mitigate causal confusion in IL, our method employs target-centric pose and ego-centric collision as outcome-oriented loss across all modalities beyond pure imitation loss.
Evaluations on real-world datasets demonstrate that MultiPark achieves state-of-the-art performance across various scenarios.
We deploy MultiPark on a production vehicle, further confirming our approach’s robustness in real-world parking environments.
The summary video and real-world testing are available on our project page: \href{https://wuyi2121.github.io/MultiPark/}{\textcolor{blue}{https://wuyi2121.github.io/MultiPark/}}.
\end{abstract}


\section{Introduction}
	
Parking, as the final critical phase of driving, consistently challenges human drivers \cite{khalid2021smart}. 
Currently, rule-based parking methods dominate autonomous parking systems \cite{zhang2018autonomous,li2021optimization}. 
However, they exhibit unstable computation time and success rates in cluttered environments \cite{zhou2020dl}. 
Moreover, they depend on modular autonomous parking systems, accumulating errors across modules and critically relying on high-precision perception, localization, and mapping \cite{zheng2023fast}.
Most importantly, overly simplified environmental modeling (e.g., convex polygon obstacles) constrains real-world applicability \cite{song2022time}.
Therefore, it's necessary to adapt end-to-end learning paradigms to parking for achieving lossless information flow and joint global optimization.

\begin{figure}[t]
	\centering
	\subfigure[Workflow of MultiPark.]{
		\includegraphics[width=.46\linewidth]{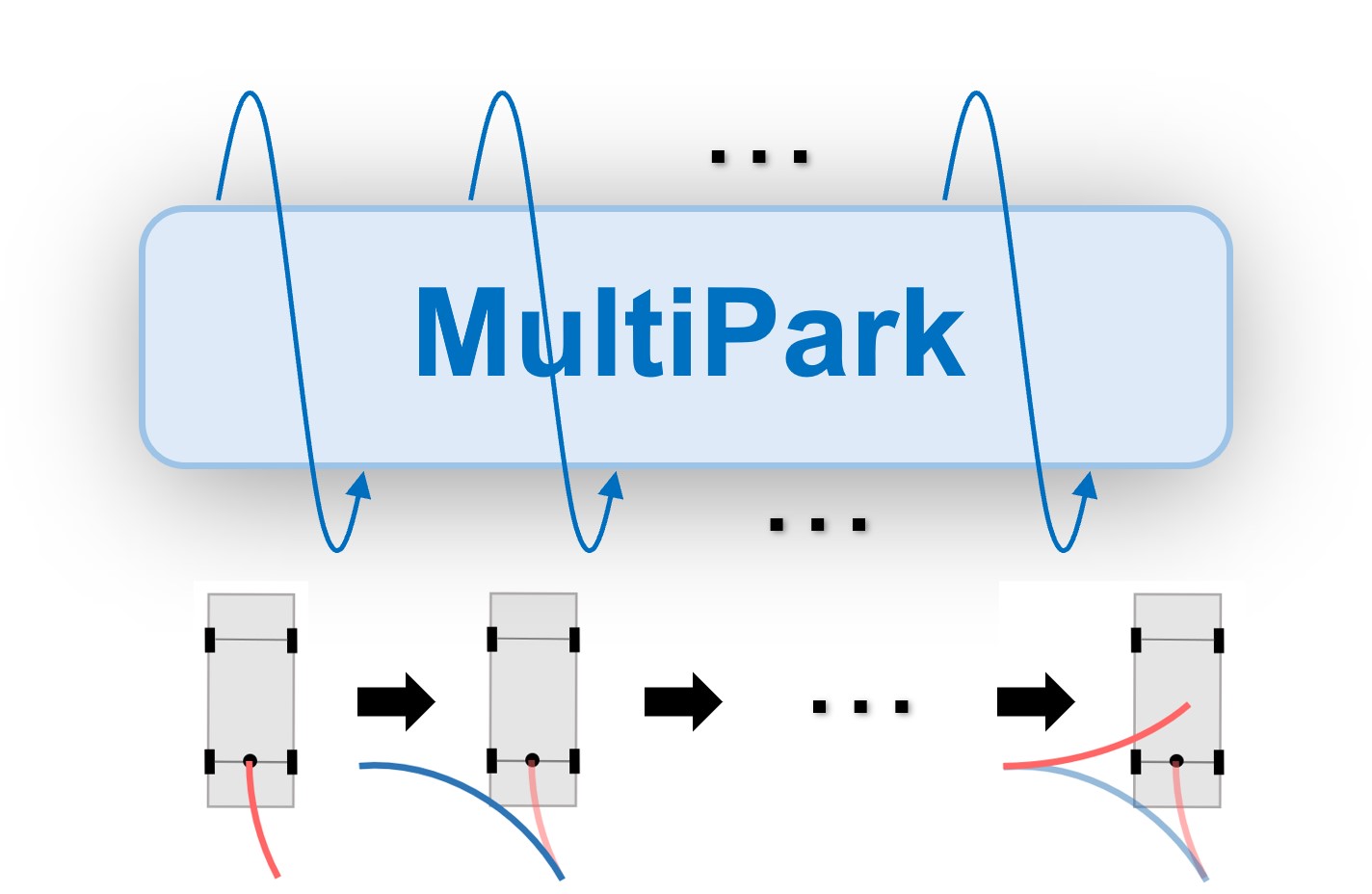} \label{fig:MultiPark}
	}
    \subfigure[Discontinuous segments.]{
		\includegraphics[width=.45\linewidth]{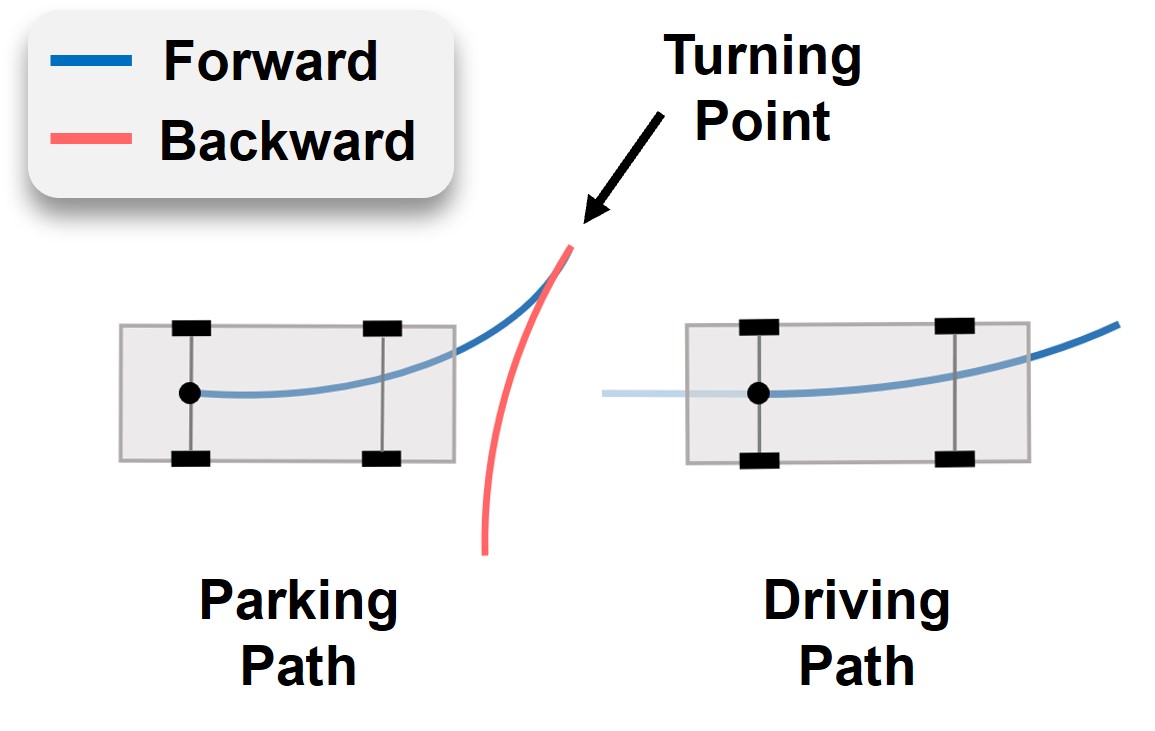} \label{fig:motivation_1}
	}
	\subfigure[Multimodal parking paths.]{
		\includegraphics[width=.46\linewidth]{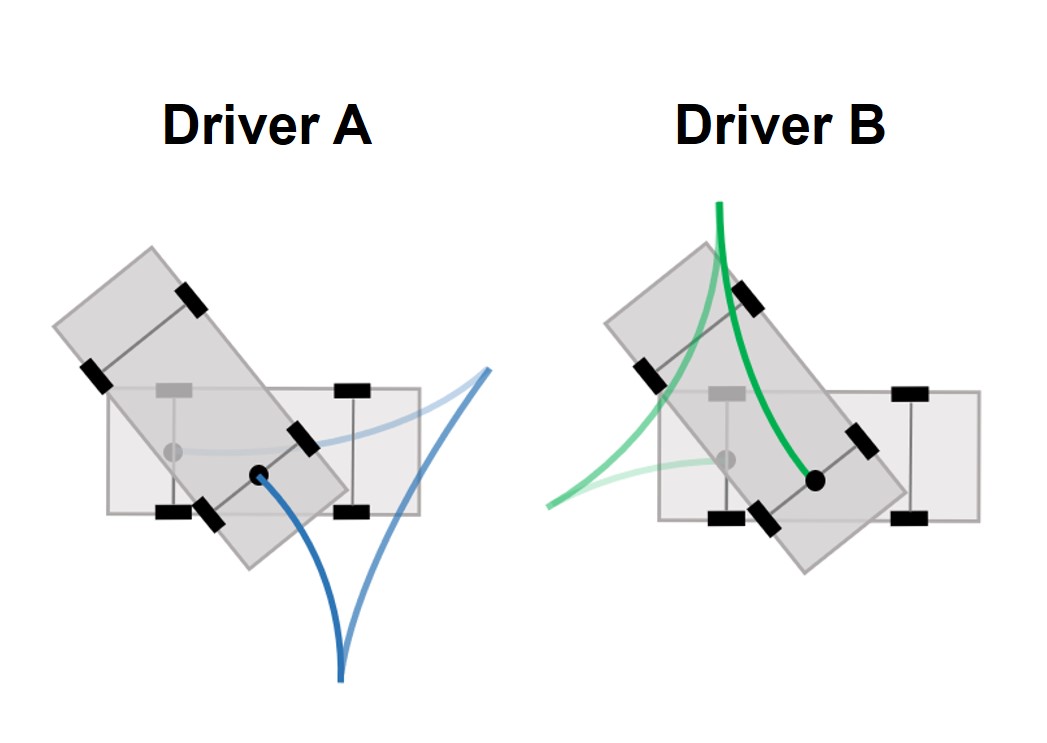} 
		\label{fig:motivation_2}
 	}
        \subfigure[Distribution shifts.]{
		\includegraphics[width=.46\linewidth]{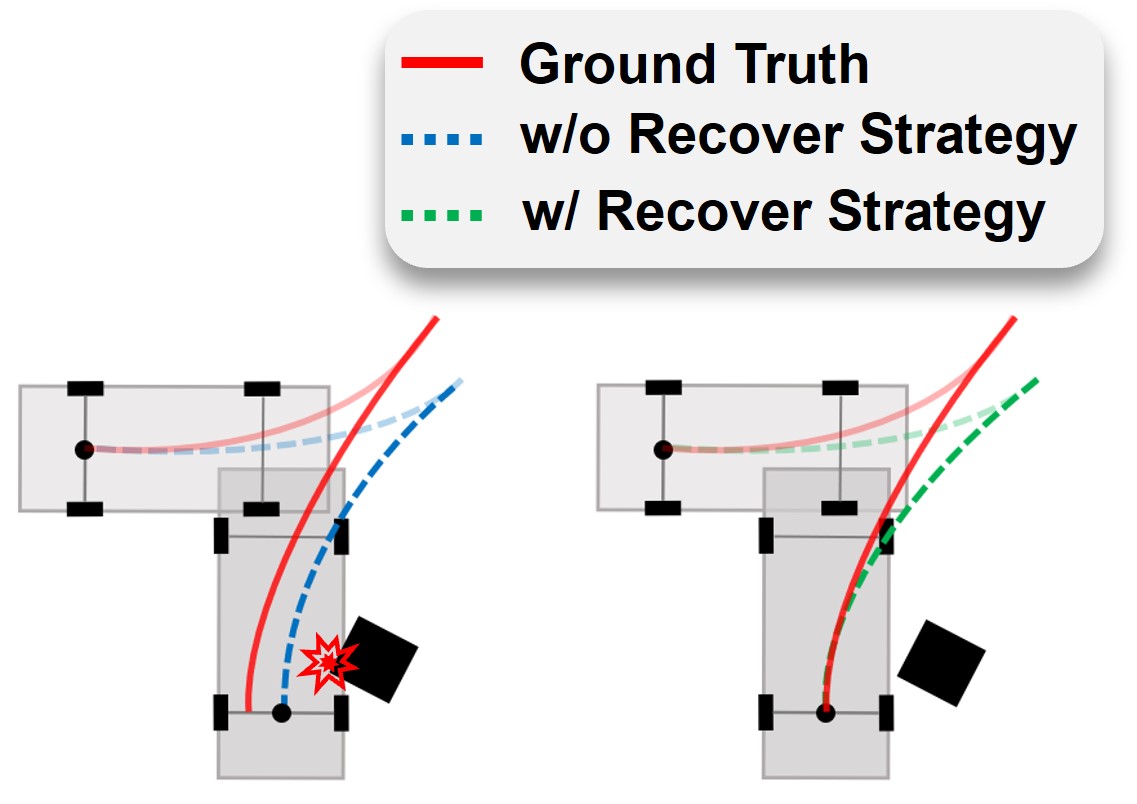} 
		\label{fig:motivation_3}
 	}
    \captionsetup{font=footnotesize} 
	\caption{In (a), we propose MultiPark that utilizes next-segment prediction for multimodal parking.
    In (b), the parking path involves both forward and backward gears, resulting in discontinuous segments with sharp turning points. In (c), under the same scenario, different drivers may yield distinct parking solutions. In (d), the blue path accumulates errors and causes collisions, but the green path can recover from mistakes occurring in the first segment.
    } 
	\label{fig:challenge}
        \vspace{-30pt}
\end{figure}

Although end-to-end autonomous driving has become mainstream in recent years \cite{tampuu2020survey,chen2024end}, there is still a long way to achieve robust end-to-end parking from the following three perspectives. 
\textit{First}, unlike driving paths exhibiting smooth continuity with a constant motion direction, parking paths contain discontinuous segments with sharp turning points, which reduces data utilization efficiency and requires modeling cross-segment causal dependencies, as shown in Fig. \ref{fig:motivation_1}.
\textit{Second}, predicting multimodal parking paths is essential since
the training data distribution for parking is inherently multimodal, as shown in Fig. \ref{fig:motivation_2}. 
Forcing the model to plan a single path to fit this distribution may lead to mode collapse, i.e., failing to capture the full distribution of the training data and producing an averaged solution that is practically meaningless \cite{li2024parkinge2e}.
\textit{Third}, learning-based parking methods inherently suffer from distribution shifts \cite{zhang2019bridging}, primarily because these models are myopically optimized for next-step action prediction rather than successful completion of long-horizon parking tasks.
Therefore, learning recovery strategies for out-of-distribution (OOD) parking scenarios becomes critically important, as shown in Fig. \ref{fig:motivation_3}.

To address these problems, we proposed MultiPark, and our contributions include: 
(i)
We pioneer a data-efficient next-segment prediction
paradigm for parking, enabling spatial generalization and temporal extrapolation.
(ii)
We design learnable parking queries factorized into gear, longitudinal, and lateral components, decoding multimodal parking behaviors in parallel.
(iii)
We adopt a two-stage training strategy with target-centric pose and ego-centric Euclidean Signed Distance Field (ESDF) collision as outcome-oriented loss across all modalities, mitigating causal confusion inherent in imitation learning.
(iv)
We achieve remarkable performance in real-world datasets, outperforming prior works.
We deploy MultiPark on
a production vehicle, verifying its generalization and effectiveness in real-world garages.

\section{Related Work}
\label{sec:Related Work}

\subsection{End-to-End Autonomous Driving.}
The transformer architecture, originally developed for sequence modeling in NLP via attention mechanisms \cite{vaswani2017attention}, has been successfully adapted to end-to-end autonomous driving. 
Specifically, TransFuser \cite{chitta2022transfuser} and InterFuser \cite{shao2023safety} employ transformer encoders to align heterogeneous camera and LiDAR features, enabling spatial-semantic consistency.
Furthermore, UniAD \cite{hu2023planning} establishes a framework integrating transformer-based perception, prediction, and planning, where full-stack driving tasks are communicated with unified query interfaces.
Unlike previous practices using deterministic modeling for planning, VADv2 \cite{chen2024vadv2} and PLUTO \cite{cheng2024pluto} utilize a transformer decoder for probabilistic modeling of multimodal behavior.
However, these imitation learning methods encompass inherent limitations, including the propensity for distribution shift and causal confusion \cite{bansal2018chauffeurnet} issues.

\subsection{Multimodal Trajectory Prediction.} 
Multimodal trajectory prediction for recovering diverse agent behaviors has been extensively studied \cite{huang2022survey}, primarily developing anchor-based and anchor-free paradigms. 
Notably, anchor-based methods represented by MultiPath \cite{chai2019multipath} leverage Gaussian mixture models with pre-defined anchors for waypoint regression, while goal-oriented variants (TNT \cite{zhao2021tnt}, DenseTNT \cite{gu2021densetnt}) decompose prediction into goal selection and trajectory generation. 
However, these methods confront inherent trade-offs between sparse anchor-induced performance degradation and dense anchor-driven expensive computation.
Conversely, anchor-free methods \cite{zhou2022hivt,tang2024hpnet} directly regress trajectories but suffer from mode collapse and training instability.
To address these limitations, MTR \cite{shi2022motion} introduces learnable motion query pairs for joint optimization of global intention localization and local movement refinement, whereas QCNet \cite{zhou2023query} combines anchor-free proposal and anchor-based refinement for trajectory decoding.

\subsection{Learning-Based Autonomous Parking.} 
Learning-based approaches achieved
promising results for handling autonomous parking recently \cite{yang2024e2e,li2024parkinge2e}.
ParkingE2E \cite{li2024parkinge2e} is the first open-source end-to-end parking network, 
discretizing continuous coordinates into tokens for sequence prediction. 
Considering a single decoder misaligns with temporal coordinate
inputs, a dual-decoder transformer \cite{du2025transparking} is used to achieve trajectory decoding.
However, the above IL-based works are limited to single-segment parking within simple scenarios.
Reinforcement learning (RL) shows promise for multi-segment planning in narrow environments recently \cite{zheng2025embodied}. 
HOPE combines RL with Reeds-Shepp curves to provide positive samples \cite{jiang2025hope}, enabling efficient learning in early training phases. 
However, RL policies suffer from persistent challenges, including significant sim2real gaps \cite{wagenmaker2024overcoming}, and low utilization of samples \cite{zhang2021sample}.
Furthermore, the human alignment problem is another challenge. 
The exploration process in RL can lead to policies that deviate from human-parking behavior\cite{gao2025rad}, such as excessive gear shifts, steering shaking, and even extreme behaviors.

\section{Method}
\label{sec:Method}
\subsection{Problem Formulation}
This work explores learning-based multimodal parking in garages. 
The model takes as input a sequence of $t$ historical surround-view camera images $\vphantom{s_j^P}\smash{\mathcal{I}_{T-t:T}={\{ \mathcal{I}_{T-t:T}^k\}}_{k=1}^{N_{cam}}}$  at timestamp $T$ and the pose $\mathcal{S} = [x_0, y_0, \psi_0]^{\top}$ of the target parking slot, where $(x_0, y_0)$ and $\psi_0$ denote the 2D coordinate and the orientation, respectively. 
Our proposed MultiPark $f_\phi: \{\mathcal{I}_{T-t:T}, \mathcal{S} \} \rightarrow \mathcal{P}_{pred}$ is a neural network parameterized by $\phi$, which generates $N_{g}$ multimodal paths $\mathcal{P}_{pred} = \{\mathbf{p}, v\}\in \mathbb{R}^{N_{s} \times N_{p} \times 4}$ associated with  scores. 
Each path $\mathcal{P}_{pred}$ includes $N_{s}$ segments with $N_{p}$ equidistant waypoints, with $\mathbf{p} = [x, y,\psi]^\top$ representing the 2D pose in the initial ego frame and $v$ indicating whether the corresponding segment is valid.
The training dataset is defined as $\mathcal{D} = \{({\mathcal{I}_{T-t:T}}_{i}, \mathcal{S}_{i,j}, {\mathcal{P}_{e}}_{i,j})\}$, where $i\in[1,N_{sce}]$ and $j\in[1,N_{slot}]$ index scenarios and target parking slots, respectively, and $\mathcal{P}_{e}$ denotes expert parking paths. 
Note that only one ground-truth (GT) path is available per training sample, but our model is tasked with deriving multiple plausible solutions to account for multimodal parking behavior. 
We summarize the optimization objective of our model as follows:
\begin{equation}
\phi^* = \arg\min_{\phi} \mathbb{E}_{(\mathcal{I}_{T-t:T}, \mathcal{S}, \mathcal{P}_{e}) \sim \mathcal{D}} \left[\mathcal{L}\left(\mathcal{P}_{e}, f_\phi(\{\mathcal{I}_{T-t:T}, \mathcal{S})\right)\right]\,,
\end{equation}
where we denote by $\mathcal{L}(\cdot, \cdot)$ the total loss function.

\subsection{The MultiPark Architecture}
\begin{figure*}[t]
	\centering
    \includegraphics[width=0.95\textwidth]{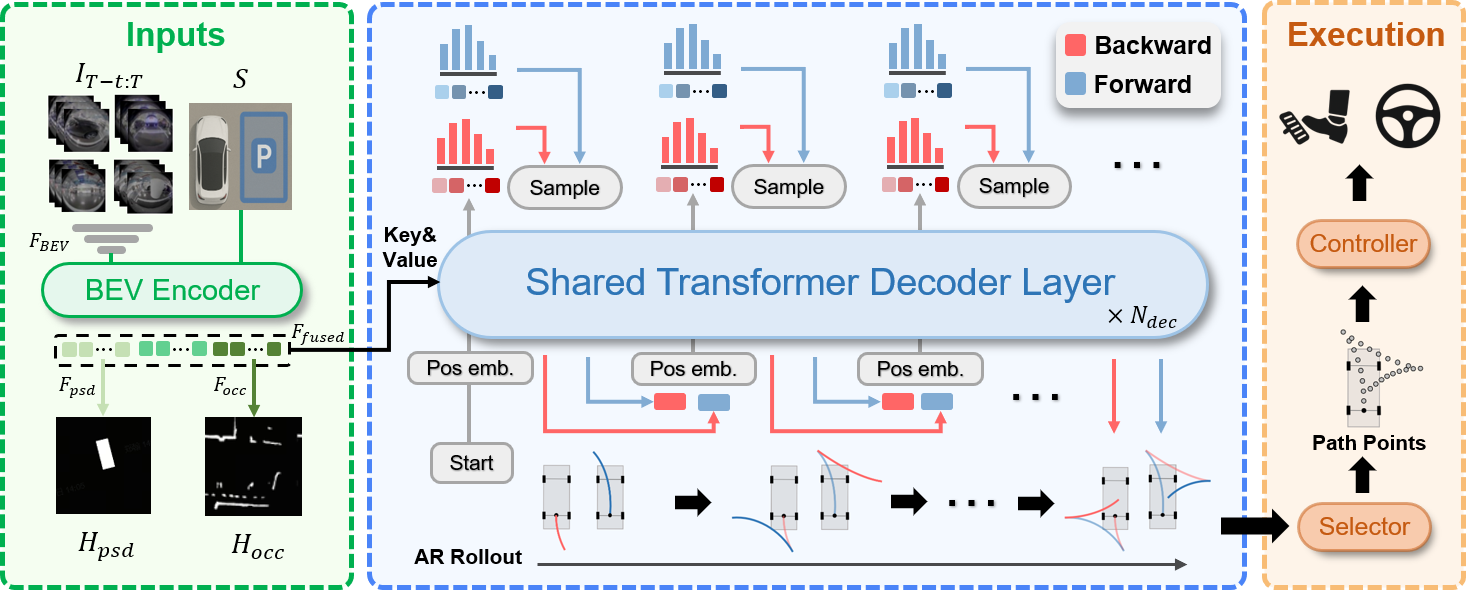}
    \captionsetup{font=footnotesize}
	\caption{Overview of MultiPark. 
    A BEV encoder first takes inputs and obtains the fused scene features as key and value for cross-attention. 
    Next, a query-based decoder parallelly predicts paths with associated scores in backward and forward gears.
    We sample from two gear-specific distributions and obtain two segments. 
    Then, reverse embedding of GSP states is utilized to achieve gear switch, thereby rolling out the next segment autoregressively and finally deriving the multimodal parking paths.
    Finally, we select the optimal path and utilize the predicted waypoints to control the vehicle.
    }
\label{fig:framework}
\vspace{-25pt}
\end{figure*}

\noindent\textbf{Overview.}
Fig.~\ref{fig:framework} depicts the overall framework of our proposed Multipark, integrated with BEV-based encoding and transformer-based decoding. 
We first introduce our encoder network for scene feature modeling in a unified BEV representation.
Given the extracted features, we present a next-segment prediction paradigm to reformulate the multimodal parking problem as an autoregressive prediction of next segment.
Finally, we propose a DETR-like transformer decoder with semi-anchor parking queries for predicting multimodal parking paths.

\noindent\textbf{Unified BEV Representation.}
Sequential RGB images $\mathcal{I}_{T-t:T}$ are first encoded by EfficientNet \cite{tan2019efficientnet} into camera features $\mathcal{F}_{cam} \in\mathbb{R}^{{C}\times{H_{cam}}\times{W_{cam}}}$. 
Through inverse perspective mapping (IPM) conditioned on intrinsics and extrinsics of the cameras \cite{qin2020avp}, we obtain a feature map
$\mathcal{F}_{BEV}$ in the BEV space. 
Then, a BEV encoder processes BEV features $\mathcal{F}_{BEV}$ to learn instance-level, temporal features of parking slot detection (PSD) $\mathcal{F}_{psd}$ and occupancy (OCC) $\mathcal{F}_{occ}$. 
Furthermore, the detection heads output explainable heatmaps $\mathcal{H}_{psd},\mathcal{H}_{occ} \in [0,1]^{H_{bev} \times W_{bev}}$ aligned to the rear axle center, where user selection $\mathcal{S}$ extracts target PSD heatmap $\mathcal{H}_{psd}$ via masking.
Finally, latent features are fused into scene features $\mathcal{F}_{fused}$, serving as the encoder output for cross‑attention in the following transformer decoder. 

\noindent\textbf{Next-Segment Prediction.} 
Parking involves frequent gear shifts. As a result, a complete path $\mathcal{P}$ can be decomposed into $N_{s}$ segments, where the gear shift points (GSP) are critical for modeling. 
Inspired by the next-token prediction paradigm in language modeling \cite{vaswani2017attention}, we factorize the joint distribution of all GSP in an autoregressive (AR) manner:
{\setlength{\abovedisplayskip}{1pt} 
\setlength{\belowdisplayskip}{1pt} 
\begin{equation}
p(\mathcal{P}|s_0,\mathcal{F}_{fused}) = \prod_{j=1}^{N_{\text{s}}} p(s_j|s_{j-1}, \mathcal{F}_{fused}),
\end{equation}
}Here, $s_j=\mathbf{p}_{j}$ represents the state of GSP with segment index $j\in[0,N_{s}]$, $s_0=[0,0,0]$ denotes the first GSP state, i.e., the initial pose in the ego frame, and we assume the Markov property $p(s_j|s_{1:j-1}, \mathcal{F}_{fused}) = p(s_j|s_{j-1}, \mathcal{F}_{fused})$ holds for static parking scenarios. 
Given the GSP state $s_{j-1}$ in the $(j-1)$-th segment and the fused scene features $\mathcal{F}_{fused}$, we can predict the next GSP state $s_{j}$ autoregressively, thereby completing the prediction of all GSP states $\vphantom{s_j^P}\smash{{\{ s_j\}}_{j=1}^{N_{s}}}$.
To fine-grain the $j$-th segment and reach the next GSP state $s_{j}$, we task our model with predicting a curvature chunk $\mathcal{C}_j=\{ \Delta s, \kappa_1,\kappa_2,...,\kappa_{N_{p}}\}$, i.e., a space-uniform $N_{p}$-piece segment parameterized by equidistant arc length $\Delta s$ and curvature sequence $\vphantom{s_j^P}\smash{{\{ \kappa_l\}}_{l=1}^{N_{p}}}$.  
Specifically, given the predicted curvature chunks $\vphantom{s_j}\smash{{ \mathcal{C}_j}\in \mathbb{R}^{N_{p}}}$, the intermediate
waypoints $\vphantom{s_j^P}\smash{\mathcal{W}_j={\{ \mathbf{p}_{j}^{k}\}}_{k=1}^{N_{p}}}$ can be integrated via:
{\setlength{\abovedisplayskip}{8pt} 
\setlength{\belowdisplayskip}{8pt} 
\begin{equation}
\mathbf{p}_{j}^{k+1} = \mathbf{p}_{j}^{k} + f_{RK2}(\mathcal{C}_j, \mathbf{p}_{j}^{k}),
\end{equation}
}where $f_{RK2}$(·) is the RK2 method \cite{rossler2009second}, $j\in[0, N_{s}]$ and $k\in[0, N_{p}]$ index segments and waypoints, respectively, $\vphantom{s_j^P}\smash{\mathbf{p}_{j}^{k}\in \mathbb{R}^3}$ denotes the $k$-th predicted waypoint in the $j$-th segment.
Given the integrated waypoints $\vphantom{s_j^P}\smash{\mathcal{W}_j}$ in the $j$-th segment, the startpoint $\vphantom{s_j^P}\smash{\mathbf{p}_{j+1}^{0}}$, i.e., the GSP state $s_j$ in the $j$-th segment can be updated, thereby enabling AR rollout.
Moreover, the model also predicts whether the segment is valid, identifying the completeness of the parking process.

\begin{figure}[t]
    \centering
    \includegraphics[width=0.38\textwidth]{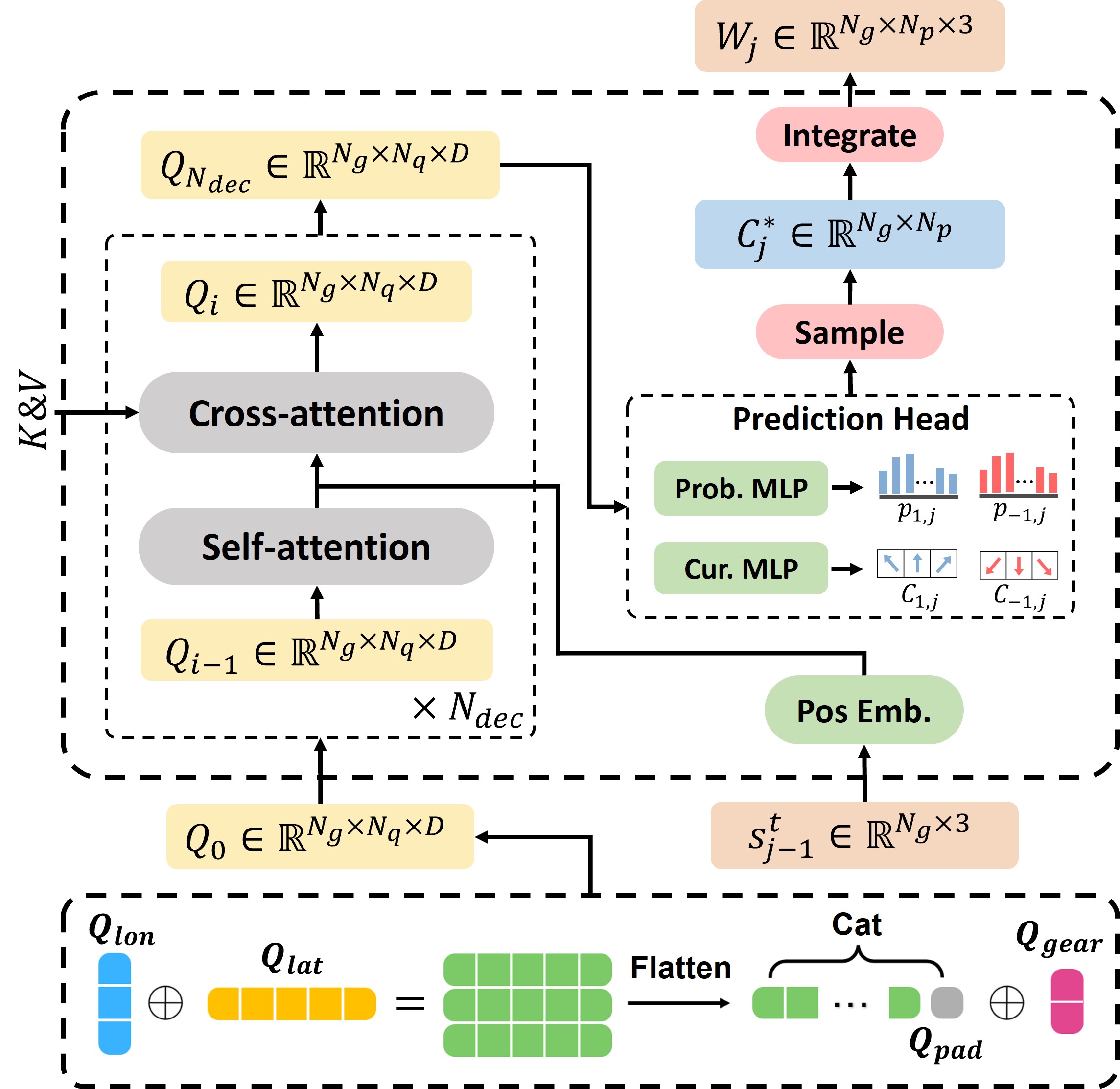}
    \captionsetup{font=footnotesize}
    \caption{The structure of our decoder network with semi-anchor parking queries in the $j$-th segment prediction.}
    \label{fig:decoder_network}
    \vspace{-15pt}
\end{figure}

\noindent\textbf{Semi-Anchor Parking Queries.} 
We utilize a DETR-like~\cite{carion2020end} decoder to predict multimodal curvature chunks in parallel. 
As shown in Fig. \ref{fig:decoder_network}, the decoder contains $N_{\text{dec}}$ stacked transformer decoder layers, each comprising parking behavior self-attention and query-to-scene cross-attention. 
Additionally, two prediction heads are appended to generate multimodal curvature chunks with associated scores.
As evidenced by~\cite{liu2021multimodal}, anchor-free queries easily result in mode collapse and training instability.
We propose semi-anchor parking queries, i.e., combining independent learnable queries representing different semantic behaviors.
Specifically, given the fact that multimodal parking behaviors can be decomposed into gear (e.g., \textit{forward} and \textit{backward}), longitudinal (e.g., \textit{short}, \textit{middle}, and \textit{long}), and lateral (e.g., \textit{sharp left}, \textit{slight left}, \textit{straight}, \textit{slight right}, and \textit{sharp right}) behaviors, a small number of gear queries $Q_{\mathrm{gear}} \in \mathbb{R}^{N_{g} \times D}$, longitudinal queries  $Q_{\mathrm{lon}} \in \mathbb{R}^{N_{lon} \times D}$, and lateral queries $Q_{\mathrm{lat}} \in \mathbb{R}^{N_{lat} \times D}$ are combined to create the initial set of parking queries concisely via
{\setlength{\abovedisplayskip}{8pt} 
\setlength{\belowdisplayskip}{8pt} 
\begin{equation}
    Q_0 = (\text{cat}(Q_{\mathrm{lon}} \oplus Q_{\mathrm{lat}}, Q_{\mathrm{pad}})) \oplus Q_{\mathrm{gear}} \in \mathbb{R}^{N_{g}\times N_{q} \times D}\,,
\end{equation}
}where $\oplus$ denotes broadcasted element-wise addition, cat(·) is the concatenation operation, $D$ denotes the hidden dimension of the embeddings, and $Q_{\mathrm{pad}}\in \mathbb{R}^{1 \times D}$ denotes the padding queries for ignoring invalid segments.
Finally, the combined queries form $Q_0 \in \mathbb{R}^{N_{g}\times N_{q} \times D}$, where $N_q = N_{lon}N_{lat} + 1$. 
Each query predicts a mode-specific parking path, stabilizing training and enabling multimodal prediction.

\noindent\textbf{Query-Based Multimodal Decoding.} 
For the $i$-th decoder layer in the $j$-th segment prediction, we first apply self-attention on the content query features $Q_{i-1}$ propagated from the $(i-1)$-th layer to explicitly model cross-query interactions, thereby enhancing the diversity of parking behaviors.
Next, we utilize cross-attention to aggregate segment-specific features with the content query features $Q'_{i-1}$ from self-attention, positional query features from the embedding of the GSP state $s_{j-1}^t$ transformed into the target slot frame, and the fused scene features $\mathcal{F}_{fused}$ as key and value. 
Specifically, two attention mechanisms are formulated as
\begin{align}
&Q'_{i-1} = \text{SelfAttn}\left(Q_{i-1}\right), \\
&Q_i = \text{CrossAttn}(Q'_{i-1}+\text{PE}(s_{j-1}^t), \mathcal{F}_{fused}),
\end{align}
where PE(·) is the position embedding, and $s_{j-1}^t$ is invariant under transformations of the frames, thereby being shared across all segments and scenarios.
The output $Q_{N_{\text{dec}}}$ from the last decoder layer is then processed by two MLP heads. 
Specifically, the regression head employs sigmoid and renormalization to directly output multimodal curvature chunks $\vphantom{s_j}\smash{{ \mathcal{C}_j}\in \mathbb{R}^{N_{g}\times N_{q}\times N_{p}}}$, while the classification head with a softmax function predicts the probability distribution $\vphantom{s_j}\smash{{p_j}\in \mathbb{R}^{N_{g}\times N_{q}}}$, with its GT generated as a one-hot distribution through anchor matching.
Finally, the optimal curvature chunks $\vphantom{s_j}\smash{{ \mathcal{C}_j^*}\in \mathbb{R}^{N_{g}\times N_{p}}}$ are sampled from two gear-specific probability distributions $\vphantom{s_j}\smash{{ p_{i,j}}\in \mathbb{R}^{N_{q}}}$, where $i\in\{-1, 1\}$ index gears.
Then, the waypoints $\vphantom{s_j}\smash{{W_{j}}\in \mathbb{R}^{N_{g}\times N_{p}\times 3}}$ can be integrated via Eq. (3) and the startpoints in the $(j+1)$-th segment can be updated by the endpoints in the $j$-th segment through $\vphantom{s_j^P=s_j^P}\smash{{\mathbf{p}_{-1,j+1}^{0} =\mathbf{p}_{1,j}^{N_{p}}}}$ and 
$\vphantom{s_j^P=s_j^P}\smash{{\mathbf{p}_{1,j+1}^{0} =\mathbf{p}_{-1,j}^{N_{p}}}}$, achieving switches between forward and backward gears in adjacent segments. 

\subsection{Two-Stage Training Strategy}
A fundamental challenge inherent in IL is how to overcome distribution shift and causal confusion \cite{filos2020can}. 
We utilize a two-stage training strategy to address the above problems.
In the pretraining stage, we adopt teacher-forcing to parallelize the modeling of next-segment prediction, reducing the learning difficulty. 
Then, we introduce argmax finetuning with outcome-oriented loss across all modalities, enabling our model to adapt to OOD scenarios and capture the causal relationships of the parking task.
To prevent degradation in the finetuning stage, the imitation loss is still added as regularization. 
The details of each stage are described below.

\noindent\textbf{Teacher-Forcing with Imitation Loss.} 
Following the typical teacher-forcing training of language models \cite{vaswani2017attention}, we parallelly predict multimodal outputs based on the GT GSP $\vphantom{s_j}\smash{s_{j-1}^{gt}}$ in each segment.
Then, we employ a winner-takes-all strategy \cite{zhou2023query}, where only the GT modality sampled from $\vphantom{s_j}\smash{p_{j}^{gt}}$ is selected for backpropagation. 
Specifically, the loss function $\mathcal{L}_{\text{teach}}$ is achieved through segment-wise summation:
\begin{equation}
\mathcal{L}_{\text{teach}} = \mathcal{L}_{\text{imitaion}} = \sum\limits_{j=1}^{N_{\text{s}}}( \underbrace{\lambda_{\text{p}}\mathcal{L}_{\text{wpt}} + \lambda_{\text{c}}\mathcal{L}_{\text{cur}}}_{\text{Regression}} + \underbrace{\lambda_{\text{p}}\mathcal{L}_{\text{prob}} + \lambda_{\text{v}}\mathcal{L}_{\text{val}}}_{\text{Classification}} ), 
\end{equation}
where the regression employs the smooth L1 loss ($\mathcal{L}_{\text{wpt}}$ and $\mathcal{L}_{\text{cur}}$ for predicting the desired waypoints and curvature chunks, respectively), while the classification utilizes the cross-entropy loss $\mathcal{L}_{\text{prob}}$ for score prediction and the binary cross-entropy loss $\mathcal{L}_{\text{valid}}$ for validity prediction, with $\lambda_{*}$ balance the four loss terms.
\begin{figure}[t]
    \centering
    \includegraphics[width=0.35\textwidth]{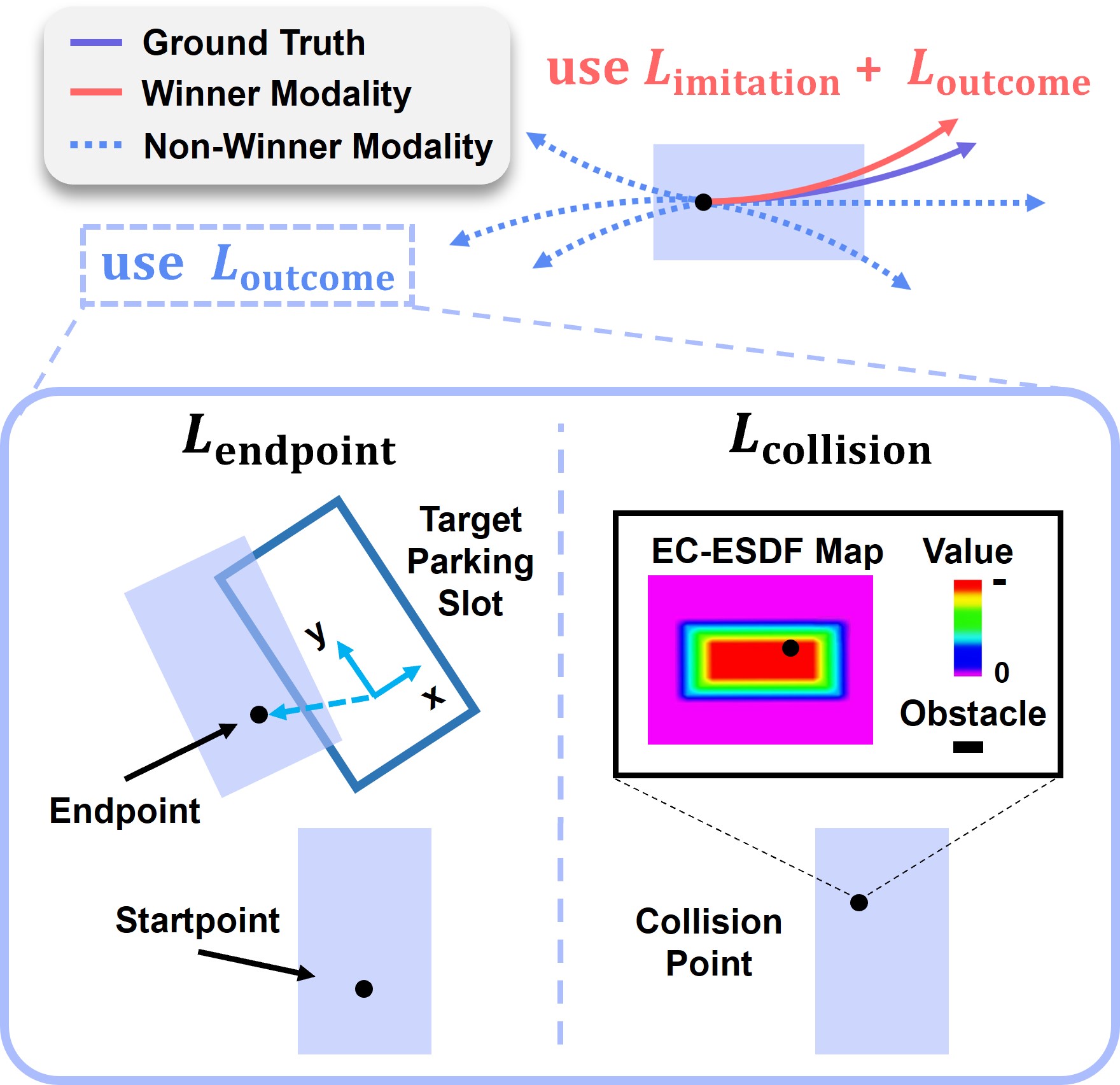}
    \captionsetup{font=footnotesize}
    \caption{The illustration of imitation loss on winner modality and outcome-oriented loss across all modalities.}
    \label{fig:outcome}
    \vspace{-25pt}
\end{figure}

\noindent\textbf{Argmax Finetuning with Outcome-Oriented Loss.}
Given a pretrained policy, we further introduce argmax finetuning, i.e., employing on-policy rollout (e.g., argmax in AR models) during training, to address distribution shifts inherent in teacher-forcing and narrow the gap between open-loop training and closed-loop deployment.
Furthermore, inspired by rewards in RL, we innovatively propose an outcome-oriented loss to enhance pure imitation loss via optimizing $\mathcal{L}_{\text{argmax}}$:
\begin{equation}
\mathcal{L}_{\text{argmax}} = \underbrace{\lambda_{\text{i}}\mathcal{L}_{\text{imitation}}(\mathcal{M}_{\text{win}})}_{\text{Regularization}} + \underbrace{\lambda_{\text{o}}\mathcal{L}_{\text{outcome}}(\mathcal{M}_{\text{all}})}_{\text{Exploration}}.
\end{equation}
where imitation loss
$\mathcal{L}_{\text{imitation}}$ for winner modality $\mathcal{M}_{\text{win}}$ serves as the regularization term to better align with human-parking behavior, outcome-oriented loss $\mathcal{L}_{\text{outcome}}$ across all modalities $\mathcal{M}_{\text{all}}$ promotes multimodal exploration, as shown in Fig. \ref{fig:outcome}, and $\lambda$ balances the objectives.
Compared with sparse reward issues in prior RL works \cite{jiang2025hope}, our method achieves stable training and rapid convergence, which benefits from teacher-forcing pretraining and imitation regularization.
Besides, instead of direct RL finetuning, outcome-oriented loss circumvents the sim2real transfer challenge while preserving RL’s strength in modeling the causations.

\begin{figure*}[t]
	\centering
	\subfigure[Remarkable performance in difficult scenes.]{
		\includegraphics[width=.48\linewidth]{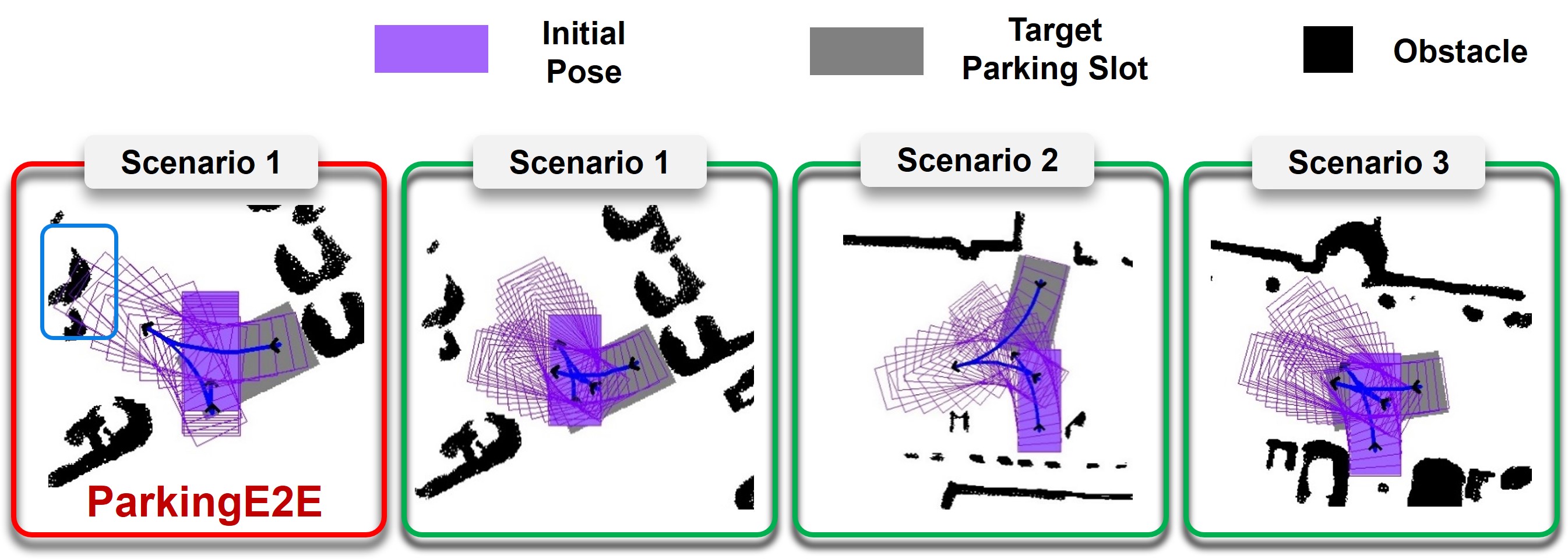} 
		\label{fig:extreme}
 	}
        \subfigure[Human-like behavior under parking queries.]{
		\includegraphics[width=.48\linewidth]{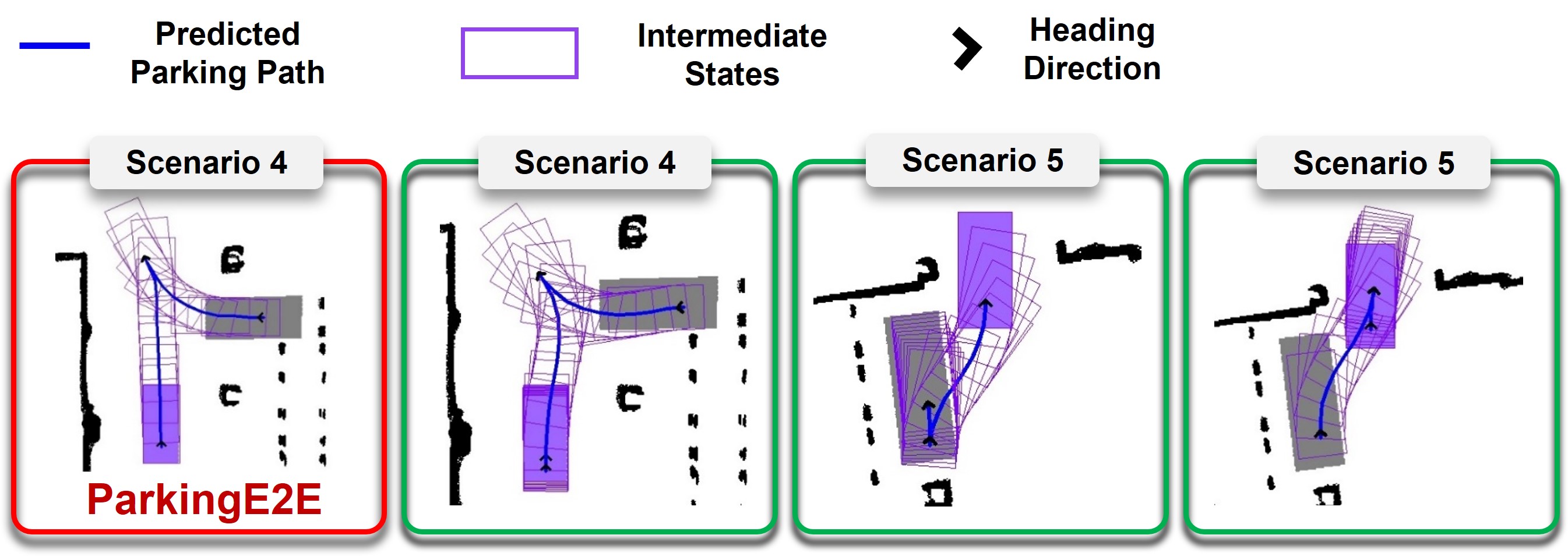} \label{fig:flexible_anchor}
	}
	\subfigure[Multimodal solutions in the same situations.]{
		\includegraphics[width=.48\linewidth]{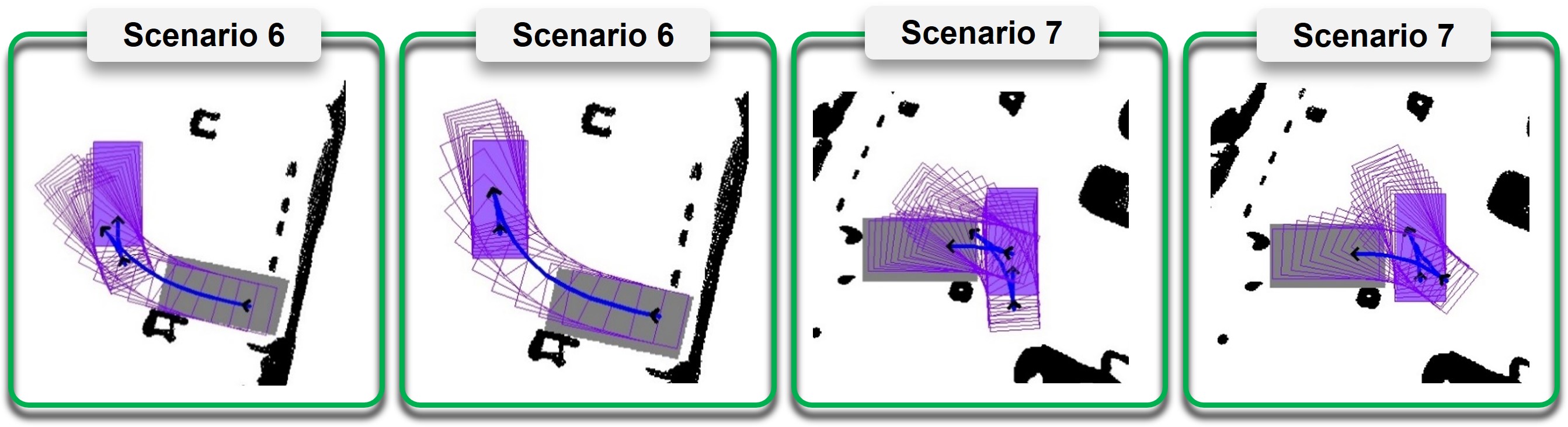} 
		\label{fig:multimodal}
 	}
    \subfigure[Without argmax loss or imitation loss.]{
		\includegraphics[width=.48\linewidth]{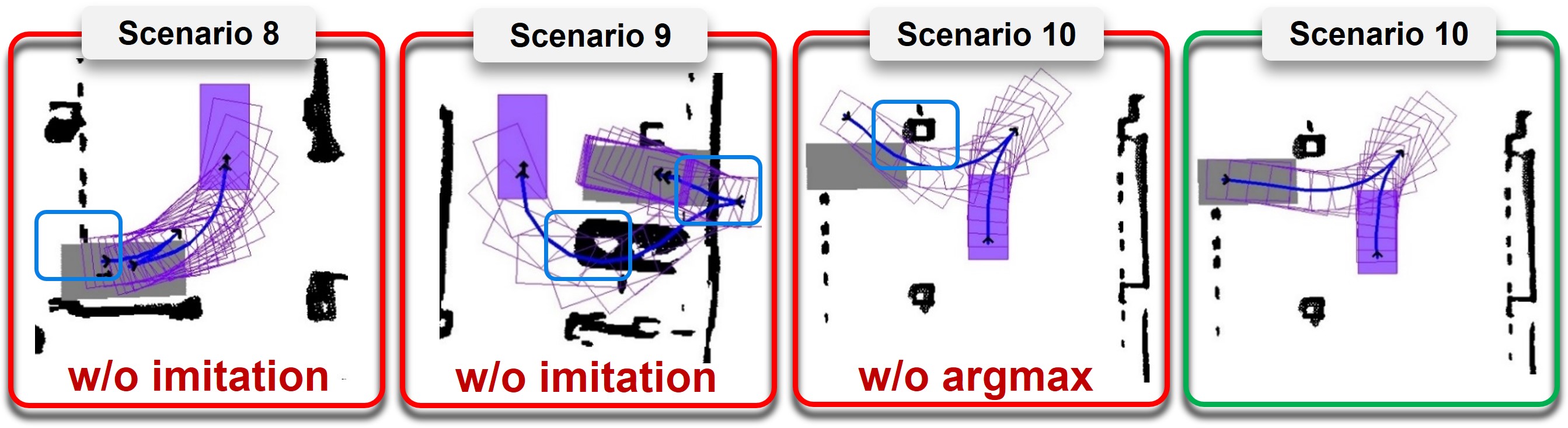} \label{fig:argmax_no_imitation}
	}
    \captionsetup{font=footnotesize} 
	\caption{Qualitative results of end-to-end parking for ten representative scenarios on the real-world datasets.
    MultiPark (marked in green) performs well in various cases, but comparative methods (marked in red) often fail.  
    The light blue frames indicate collisions happening.
}
	\label{fig:visual}
        \vspace{-10pt}
\end{figure*}

Specifically, $\mathcal{L}_{\text{outcome}}$ includes endpoint loss $\mathcal{L}_{\text{endpoint}}$ and collision loss $\mathcal{L}_{\text{collision}}$.
The primary task of parking is to match the target parking slot accurately. 
Therefore, $\mathcal{L}_{\text{endpoint}}$ is designed to minimize the target-centric endpoint loss $\mathcal{L}_{\text{endpoint}} = \|\mathbf{p}_{t}\|^2$, where $\mathbf{p}_{t}$ transforms the endpoint $\mathbf{p}_{s}$ in the startpoint frame to the target parking slot frame. 
In addition, planning collision-free paths is safety-critical. 
We adopted ego-centric ESDF (EC-ESDF) in \cite{geng2023robo} to accurately describe any-shape vehicles.
The EC-ESDF map is defined with negative interior values corresponding to the signed distance to the ego boundary and zero exterior values, ensuring continuity for differential optimization.
Specifically, the collision points $\mathbf{q}_{w}$ in the world frame are transformed to $\mathbf{q}_{e}$ in the ego frame. 
Then, values $d_{i}$ of the $i$-th collision point can be easily obtained by querying the EC-ESDF map. 
We define the differentiable collision loss as the sum of EC-ESDF value $\sum_{i=1}^{N_{\text{c}}} \left( d_{i} \right)^2$ of all collision points, where $N_{\text{c}}$ is the number of collision points. 


\section{Results}
\label{sec:Results}
We evaluate MultiPark in real-world datasets and deploy it on a production vehicle for real-world testing.
We formulate our experiments to answer the following questions:

\begin{enumerate}[label=\textbf{Q\arabic*.}, leftmargin=*, align=left]
    \item \textit{How does MultiPark compare to prior works for parking tasks?}
    \item \textit{How important is the modeling of the next-segment prediction compared to the next token prediction?}
    \item \textit{Can the semi-anchor parking queries improve both diversity and accuracy?}
    \item \textit{What is the critical factor behind the two-stage training strategy to improve the performance of MultiPark?}
\end{enumerate}

\begin{table*}[ht]
  \setlength{\tabcolsep}{4.3pt} 
  \renewcommand{\arraystretch}{1.0}
  \centering
  \begin{threeparttable}
  \caption{Performance comparison with prior works in various scenarios.}
  \begin{tabular}{@{}l*{14}{c}@{}}
    \toprule
    \multirow{1}{*}{\makecell[c]{Method}} 

    & Long. Offset(m)$\downarrow$ & Lat. Offset(m)$\downarrow$ & Orie. Offset(deg)$\downarrow$ & Cover. Rate(\%)$\uparrow$ 
    & Coll. Rate(\%)$\downarrow$ & Coll. Prop(\%)$\downarrow$ & Succ. Rate(\%)$\uparrow$ \\
    \midrule
    ParkingE2E \cite{li2024parkinge2e} & 0.2794 & 0.1540 & 10.182 & 0.9211 & 0.3167 & 0.0712 & 0.6453  \\
    TransParking \cite{du2025transparking} & 0.5901 & 0.2005 & 11.269 & 0.8816 & 0.3120 & 0.0606 & 0.5933  \\
    TransFuser \cite{chitta2022transfuser} & 0.6069 & 0.2142 & 41.476 & 0.7072 & 0.6359 & 0.1522 & 0.1134   \\
    GoalStatus \cite{li2024ego} & 0.7161 & 0.3867 & 2.4257 & 0.8661 & 0.4326 & 0.0895 & 0.4680 \\
    ParkingMLP \cite{zhai2023rethinking} & 0.1904 & 0.1330 & 1.5140 & 0.9665 & 0.2458 & 0.0511 & 0.7304 \\
    \addlinespace[1pt]
    \hline
    \addlinespace[2pt]
    OnlyReg & 0.1873 & 0.1095 & 1.7871 & 0.9564 & 0.2151 & 0.0349 & 0.7470  \\
     \rowcolor{mygreen} MultiPark & \textbf{0.1185} & \textbf{0.0777} & \textbf{1.0176} & \textbf{0.9720} & \textbf{0.1749} & \textbf{0.0268} & \textbf{0.7895}  \\
    \bottomrule
  \end{tabular}
  \centering
Please refer to \textit{Quantitative Results} and \textit{Benchmark and Metrics} for detailed implementations of baselines and descriptions of metrics, respectively.
\vspace{-25pt}
  \label{tab:total_performance}
  \end{threeparttable}
\end{table*}

\subsection{Experiment Setup}
\noindent\textbf{Dataset Collection.}
We construct the HP5 dataset, a hybrid dataset consisting of human-parking paths and expert paths for parking.
The human-parking paths are collected using vehicle-mounted devices from underground and surface parking lots, including various occupancy configurations, different lighting conditions, and hybrid slot types, as shown in Fig. \ref{fig:collected}.  
Surround-view cameras are employed to capture RGB images. 
Moreover, VINS-based techniques \cite{qin2018vins} achieve robust vehicle localization to obtain the accurate path information.
To further improve diversity and quantity, we randomly sample different target slots based on the collected data and ultilize an expert planner with privileged information to augment our dataset. 

\noindent\textbf{Benchmark and Metrics.}
Towards understanding \textbf{Q1}, we compare MultiPark to six competitive baselines.
Note that all baselines are trained on the same HP5 dataset until convergence, and we employ the same encoder network for fair comparison. 
We follow the experimental setup of ParkingE2E\cite{li2024parkinge2e} and evaluate the methods on their ability to (i) match the target parking slot accurately and (ii) reach the target parking slot safely.
Specifically, we use the closed-loop metrics below for evaluation.
Long./Lat. Offset denotes the average Euclidean distance between the GT and the final pose in the longitudinal/lateral axis, respectively, while Orie. Offset is the angular deviation.
Cover. Rate is the percentage of the final footprint area of the ego contained within the target parking slot boundaries, assessing parking accuracy.
Coll. Rate/ Prop denotes the probability of collision occurrence in all data samples/waypoints, respectively, measuring the safety of parking. 
Succ. Rate is defined as exceeding 90$\%$  in Cover. Rate without collisions.

\noindent\textbf{Implementation Details.}
We implement MultiPark and the baselines using the PyTorch framework. 
The neural network is trained in an end-to-end manner on 8 NVIDIA H20 GPUs with a batch size of 32, utilizing the AdamW optimizer.
Regarding the encoder, the size of BEV space is $800 \times 400$ with a resolution of 0.05 meters.
For the decoder, we stack 4 decoder layers, where all hidden sizes are set to 64, and each layer has 4 attention heads.
With learnable properties, only a small number of queries are required, e.g., 32 parking queries in our setting, factorized into 2 gear queries, 3 longitudinal queries, and 5 lateral queries.
Conducting k-means on limited parking data \cite{shen2022parkpredict+} yields highly redundant clusters\cite{chai2019multipath}, so we adopt a simpler way to obtain pre-defined anchors by uniformly sampling in gear/longitudinal/lateral space.
\begin{table}[htbp]
  \centering
    \setlength{\tabcolsep}{8pt} 
    \renewcommand{\arraystretch}{1.0}
    \begin{threeparttable}
      \caption{Causal confusion study for different perception features across scenarios with varying difficulty.}
      \begin{tabular}{ccccc}
        \toprule
        \multirow{2}{*}{\vspace*{-0.5ex}Method\vspace*{-0.5ex}}
        &  
        & \multicolumn{3}{c}{Coll. Rate(\%)$\downarrow$}  \\
        \cmidrule(r){3-5} 
        & & Normal & Complex & Extreme  \\
        \midrule
        GoalStatus & w/o OCC  & 0.1917 & 0.3698 & 0.4426 \\
        \addlinespace[1pt]
        \hline
        \addlinespace[2pt]
        & full model & 0.3013 & 0.3561 & 0.5901 \\
        ParkingE2E & w/o OCC & 0.3287 & 0.4246 & 0.5409 \\
        & w/o PSD & 0.7397 & 0.8082 & 0.7704 \\
        \addlinespace[1pt]
        \hline
        \addlinespace[2pt]
        & \cellcolor{mygreen}full model & \cellcolor{mygreen}\textbf{0.1643} & \cellcolor{mygreen}\textbf{0.2465} & \cellcolor{mygreen}\textbf{0.1803}\\
        MultiPark & w/o OCC & 0.2465 & 0.3972 & 0.4098 \\
        & w/o PSD & 0.2465 & 0.4109 & 0.2950 \\
        \bottomrule
      \end{tabular}
      Occupancy and Parking Slot Detection are abbreviated as OCC and PSD, respectively.
      \label{tab:perception_ablation}
    \end{threeparttable}
\end{table}
\vspace{-15pt}
\subsection{Quantitative Results}
For \textbf{Q1}, we report the test set results in Table \ref{tab:total_performance}.
Notably, MultiPark consistently outperforms the baselines in all metrics, underscoring its generalization across various scenarios.
Fig. \ref{fig:visual} presents comparative visualization results across ten representative scenarios, demonstrating MultiPark's significant advantages while indicating limitations for baseline methods.
The results discussion is primarily from four perspectives below.
\textit{First}, ParkingE2E\cite{li2024parkinge2e} and TransParking\cite{du2025transparking}, as the state-of-the-art end-to-end parking methods, explicitly model the point-to-point causal relationship through discretized tokens, whereas MultiPark groups points together and executes as one curvature chunk to predict the next segment while achieving better performance, as shown in Fig. \ref{fig:extreme}.
\textit{Second}, most end-to-end autonomous driving works prefer to adopt a GRU-based decoder, where the GRU cell takes in the hidden state and predicts the differential waypoints.
Due to few works focusing on parking, we migrated the simple but widely-used GRU-based decoder from TransFuser \cite{chitta2022transfuser}. 
However, TransFuser has large errors in the orientation, showing its poor performance in transferring to parking tasks.
The reason is that driving scenarios are predominantly straight-line, but parking tasks usually require sharp turns.
\textit{Third}, inspired by EgoStatus \cite{li2024ego}, we need to rethink whether current end-to-end parking models fully utilize perceptual information except for PSD information. 
We follow EgoStatus to implement a GoalStatus, using only PSD information as inputs. 
Eventually, we found that GoalStatus can achieve comparable results in normal scenarios in Table \ref{tab:perception_ablation}, but with the increase of difficulties, the effect decreases sharply.
This interesting phenomenon indicates that evaluating in overly simple scenes (e.g., single-segment parking in ParkingE2E) is meaningless, owing to good performance even without OCC information.
\textit{Finally}, to isolate the gain of our transformer architecture, we follow AD-MLP \cite{zhai2023rethinking} to implement ParkingMLP by replacing the autoregressive transformer with a simple MLP that directly regresses all waypoints, but retaining other components for fair comparison. 
Notably, we modify the output to predict multiple anchor-free paths simultaneously, addressing the ineffectiveness of outcome-oriented loss in single-mode planning. 
Surprisingly, this MLP-based variant outperforms ParkingE2E, demonstrating that MultiPark's remarkable performance also gains from our other designs, not solely from transformer architecture, which will be further illustrated in the following ablation studies.

\begin{table*}[ht]
  \setlength{\tabcolsep}{3pt} 
  \renewcommand{\arraystretch}{1.0}
  \centering
  \begin{minipage}{0.18\textwidth}
  \captionsetup{font=small}
  \caption{Inference time test for AR models.}
  \begin{tabular}{@{}l | *{1}{c}@{}}
    \toprule
    \multirow{1}{*}{\makecell[c]{Method}} 
    & \multicolumn{1}{c}{\makecell[c]{Infer. time \\(ms)$\downarrow$}} \\
    \midrule
    ParkingE2E  & 224.42  \\
    TransParking  & 226.94  \\
    OnlyReg  & 60.306  \\
    \rowcolor{mygreen} MultiPark  & \textbf{59.062}  \\
    \bottomrule
  \end{tabular}
  \label{tab:inference}
  \end{minipage}
\hspace{0.7cm}
\begin{minipage}{0.35\textwidth}
\captionsetup{font=small}
  \caption{Ablation study for supervising objects in outcome-oriented loss.}
  \begin{tabular}{@{}l | *{3}{c}@{}}
    \toprule
    \multirow{1}{*}{\makecell[c]{Supervise}} 
    & \multicolumn{1}{c}{\makecell[c]{Long. Offset\\(m)$\downarrow$}} 
    & \multicolumn{1}{c}{\makecell[c]{Lat. Offset\\(m)$\downarrow$}} 
    & \multicolumn{1}{c}{\makecell[c]{Orie. Offset\\(deg)$\downarrow$}} \\
    \midrule
    no modality  & 0.1871 & 0.1083 & 1.7503 \\
    winner  & 0.1303 & 0.0879 & 1.2014 \\
    non-winner  & 0.1135 & \textbf{0.0752} & \textbf{0.7660} \\
    \rowcolor{mygreen}all modality  & \textbf{0.1085} & 0.0800 & 0.8913 \\
    \bottomrule
\end{tabular}
    \label{tab:outcome}
    \end{minipage}
\hspace{0.8cm}
  \begin{minipage}{0.32\textwidth}
  \captionsetup{font=small}
  \caption{Ablation study for loss components in two-stage training strategy.}
  \begin{tabular}{@{}l | *{3}{c}@{}}
    \toprule
    \multirow{1}{*}{\makecell[c]{Loss}} 
    & \multicolumn{1}{c}{\makecell[c]{Cover. Rate\\(\%)$\uparrow$}} 
    & \multicolumn{1}{c}{\makecell[c]{Coll. Rate\\(\%)$\downarrow$}} 
    & \multicolumn{1}{c}{\makecell[c]{Succ. Rate\\(\%)$\uparrow$}}\\
    \midrule
    w/o imitation  & 0.8882 & 0.2198 & 0.6595 \\
    w/o endpoint  & 0.8508 & 0.2293 & 0.5059 \\
    w/o argmax  & 0.7541 & 0.3215 & 0.5082 \\
    \rowcolor{mygreen}full training  & \textbf{0.9720} & \textbf{0.1749} & \textbf{0.7895} \\
    \bottomrule
  \end{tabular}
  \label{tab:imitation}
  \end{minipage}
\end{table*}

\subsection{Ablation Studies}
For \textbf{Q2}, we compare our MultiPark to next-token prediction baselines (e.g., ParkingE2E). 
As shown in Table \ref{tab:inference}, MultiPark significantly reduces the number of AR rollouts and achieves 4x faster inference, contributing to real-time deployment in production vehicles. 
Additionally, Fig. \ref{fig:flexible_anchor} demonstrates that MultiPark predicts smooth, human-like paths, while ParkingE2E exposes unnatural behaviors as it directly discretizes coordinates to tokens.
Table \ref{tab:perception_ablation} shows that blanking OCC leads to failed perception but minor degradation, even slight improvements in ParkingE2E's collision performance, revealing its causal confusion.

For \textbf{Q3}, we aim to verify our hypothesis that modeling multimodal parking behaviors is important.
We modified the MultiPark to OnlyReg by removing the classification head and regressing a single path. 
The results in Table \ref{tab:total_performance} and Table \ref{tab:inference} show that parallel decoding multimodalities not only achieve a comparable inference speed to OnlyReg, but also
improve all metrics by multi-modal collaboration.
Fig. \ref{fig:multimodal} illustrates that our multimodal decoder captures the full distribution of parking paths, thereby successfully predicting multiple plausible solutions under the same inputs.

For \textbf{Q4}, we conduct detailed ablation studies on supervising objects and loss components.
Table \ref{tab:outcome} demonstrates that outcome-oriented loss on winner modality yields limited improvements, likely due to merely rescaling imitation loss, while non-winner supervision significantly enhances the parking accuracy.
As shown in Fig. \ref{fig:argmax_no_imitation} and Table \ref{tab:imitation}, dropping imitation loss fails to converge like sparse reward and even generates odd behavior, proving the necessity of imitation regularization.
Moreover, argmax finetuning helps the model to recover from mistakes, as seen in Fig. \ref{fig:argmax_no_imitation}.
\subsection{Onboard Deployment and Testing}
\begin{figure}[t]
	\centering
	\subfigure[Parking lots for collecting data.]{
		\includegraphics[width=.46\linewidth]{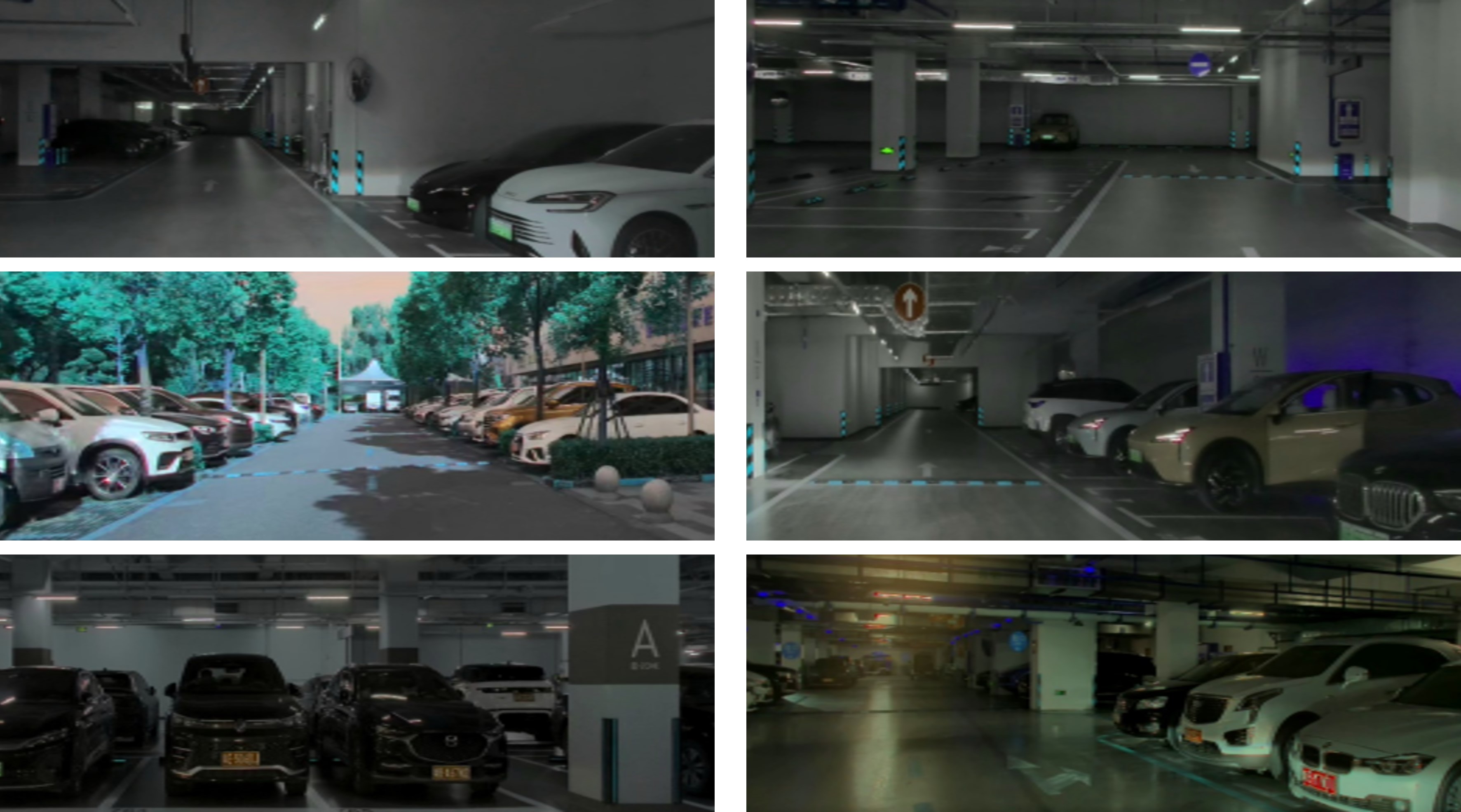} \label{fig:collected}
	}
	\subfigure[Real vehicle deployment.]{
		\includegraphics[width=.46\linewidth]{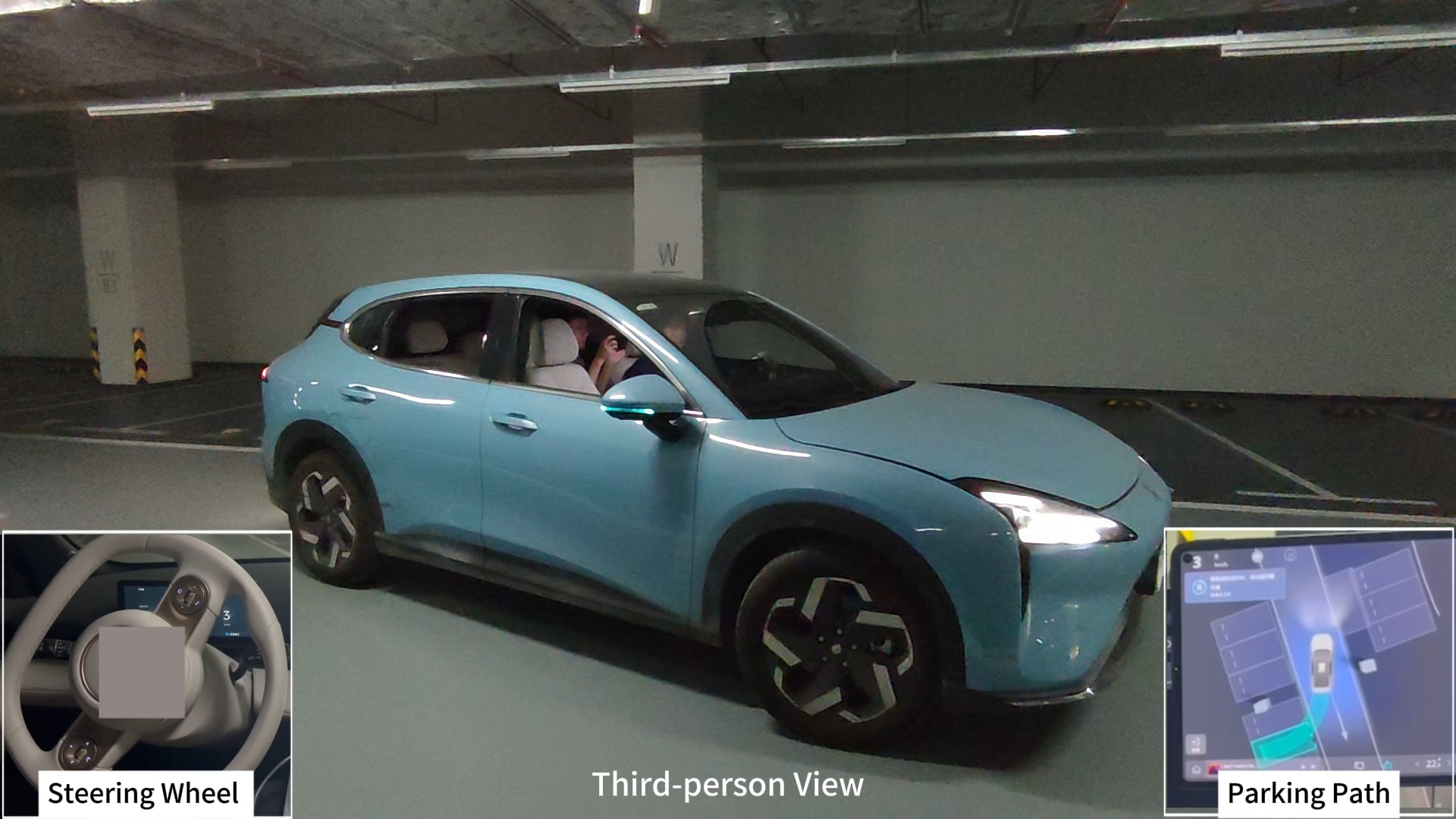} 
		\label{fig:real_vehicle}
 	}
        \subfigure[RGB images from surround-view cameras.]{
		\includegraphics[width=0.965\linewidth]{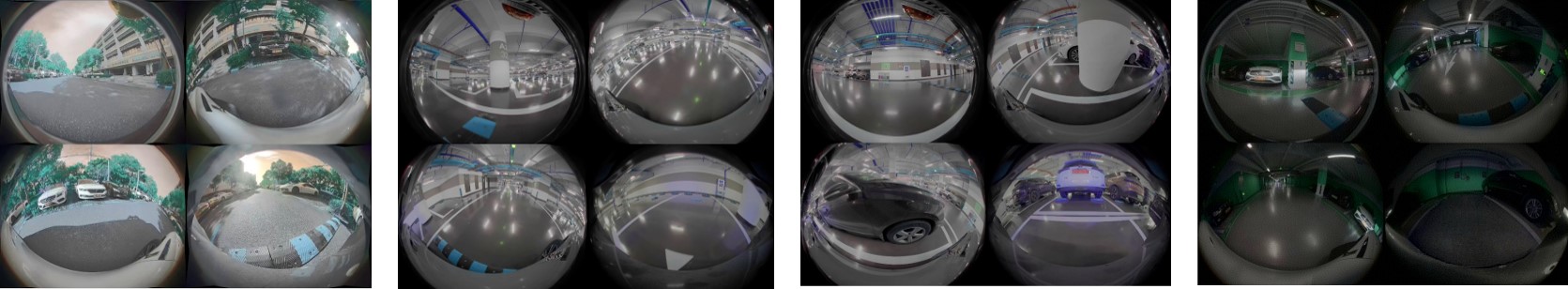} 
		\label{fig:camera}
 	}
    \captionsetup{font=footnotesize} 
	\caption{In (a), we collected data from various parking lots to enhance the diversity of training data. In (b), we deploy MultiPark on a production vehicle as the experimental platform for testing. In (c), each camera is equipped with a fisheye lens, mounted at the front, rear, left, and right sides, respectively.} 
	\label{fig:real_world}
        \vspace{-10pt}
\end{figure}

\noindent\textbf{Experimental Platform Setup.} 
Our experimental platform is a production vehicle, as illustrated in Fig. \ref{fig:real_vehicle}.
The system employs four surround-view cameras, a common setup for commercial vehicles. 
As shown in Fig. \ref{fig:camera}, each camera captures RGB images in real-time, where the intrinsic/extrinsic parameters are calibrated offline.
Furthermore, we optimize the original float network and replace the unsupported operators used by the standard transformer decoder with convolutions during ONNX export for simple deployment.
Finally, MultiPark achieves real-time path replanning on a QC8650 processor, which can handle dynamic scenarios owing to the low-speed nature of the parking scenario (e.g., pedestrians $<5km/h$, vehicles $<15km/h$).

\noindent\textbf{Onboard Parking System.}
The primary difficulty for parking lies in path planning, while velocity assignment is more tractable. 
Therefore, the onboard parking system decouples them and follows a three-stage pipeline.
\textit{First}, it plans candidate paths from RGB images and a selected slot. 
\textit{Second}, it selects the optimal path by evaluating safety and quality. 
\textit{Third}, it assigns velocity to avoid spatiotemporal overlap with dynamic obstacles (e.g., yielding to pedestrians) and tracks the optimal path via Model Predictive Control.
MultiPark completely replaces the stock parking planner, directly impacting the vehicle’s maneuver without any rule-based parking algorithm intervention in the first stage. 
We regard the second and third stages as necessary post-processing steps to inject human preference into the black-box.

\noindent\textbf{Real-World Testing.}
We conducted real-world experiments in various parking slots, as shown in Fig. \ref{fig:real_vehicle}.
Following the parking scenario difficulty standards ISO 20900 and GB/T 41630-2022 in \cite{jiang2025hope}, we categorize the testing scenarios into three difficulty levels (normal, complex, and extreme) based on the obstacle density and the size of the parking space, demonstrating our algorithm's robustness across diverse garages.
The video demonstration of challenging real-world experiments can be found in our supplementary materials.

\section{Conclusion}
We present MultiPark, an autoregressive transformer for multimodal parking.
We first introduce a data-efficient next-segment prediction paradigm for spatial generalization and temporal extrapolation.
Then, we utilize a small set of learnable parking queries for decoding multimodal parking behaviors. 
Besides, we adopt a two-stage training strategy and outcome-oriented loss to push the limit of IL in parking tasks.
Comparative results on the real-world dataset demonstrate our state-of-the-art performance, and challenging real-world experiments successfully verify our approach's generalization ability and robustness. 
We believe MultiPark offers a novel solution for building efficient parking networks.
Our future work includes leveraging the remarkable scene understanding and reasoning capability of Vision-Language Models to extend MultiPark as ParkAgent.

\bibliographystyle{IEEEtran}
\bibliography{IEEEabrv, example}  

\end{document}